\documentclass[letterpaper, 10 pt, conference]{ieeeconf}  

\IEEEoverridecommandlockouts                              

\usepackage{amsmath} 
\usepackage{makecell}
\usepackage{multirow}
\usepackage{booktabs}
\usepackage{graphicx}
\usepackage{amssymb}
\title{\LARGE \bf
Learning Multi-Skill Legged Locomotion Using Conditional Adversarial Motion Priors
}

\author{Ning Huang$^{1}$, Zhentao Xie$^{1}$, and Qinchuan Li$^{1}$
\thanks{This work supported by the Natural Science Foundation of Zhejiang province, China (No. LD25E050001). (Ning Huang and Zhentao Xie contributed equally to this work.) (Corresponding Author: Qinchuan Li.)) }
\thanks{$^{1}$School of Mechanical Engineering, Zhejiang Sci-Tech University, Hangzhou, 310018, China.}%
\thanks{$^{2}$Email:
        {\tt\small lqchuan@zstu.edu.cn.}}%
}

\begin{document}

\maketitle
\thispagestyle{empty}
\pagestyle{empty}

\begin{abstract}

Despite growing interest in developing legged robots that emulate biological locomotion for agile navigation of complex environments, acquiring a diverse repertoire of skills remains a fundamental challenge in robotics. Existing methods can learn motion behaviors from expert data, but they often fail to acquire multiple locomotion skills through a single policy and lack smooth skill transitions. We propose a multi-skill learning framework based on Conditional Adversarial Motion Priors (CAMP), with the aim of enabling quadruped robots to efficiently acquire a diverse set of locomotion skills from expert demonstrations. Precise skill reconstruction is achieved through a novel skill discriminator and skill-conditioned reward design. The overall framework supports the active control and reuse of multiple skills, providing a practical solution for learning generalizable policies in complex environments.

\end{abstract}

\section{INTRODUCTION}

In recent years, reinforcement learning (RL) has demonstrated significant potential in enabling robots to master diverse skills. In the domain of quadrupedal locomotion, RL has been successfully applied to allow quadruped robots to perform diverse and challenging tasks in real-world environments [1]–[4]. Typically, RL techniques rely on carefully designed reward functions tailored to specific tasks in order to guide the learning of desired behaviors. Yet engineering rewards for 
complex skills necessitates balancing competing objectives an error-prone process where desired behaviors seldom emerge.

Among various reinforcement learning approaches, imitation-based reinforcement learning offers a convenient and effective means for acquiring complex skills. By leveraging limited number of expert demonstrations, imitation learning enables agents to replicate expert-like behaviors. Escontrela et 
al.[5] have demonstrated that Adversarial Motion Priors (AMP) constitute an effective framework for this purpose. AMP employs a Generative Adversarial Network (GAN) structure, in which a discriminator distinguishes demonstration samples and agent-generated samples, thus providing 
reward signals that motivate the agent to produce motion behaviors stylistically similar to expert demonstrations.

Beyond simply mimicking expert motions, AMP also supports more flexible and generalizable behavior synthesis. It enables quadruped robots to learn user-specific gait patterns by enforcing motion styles specified by reference trajectories, even in the absence of direct low-level control data. This makes AMP particularly suitable for solving well-defined tasks where stylistic fidelity is important, while alleviating the need for dense or fine-grained expert annotations. However, 
the training process of GANs is known to be highly unstable[12],[15]. When learning multiple motion skills simultaneously, the generator may collapse to producing only a limited subset of behaviors, failing to cover the full data distribution. This lack of diversity in the generated samples undermines the representational richness required to capture the complexity of real expert demonstrations.

In this study, we propose a novel approach based on the fundamental idea of Conditional Generative Adversarial Networks (CGAN) to address the challenge of multi-skill learning in the Adversarial Motion Priors (AMP) framework. Traditional AMP frameworks usually focus on learning a single target task and cannot effectively handle complex multi-skill learning problems. To overcome this limitation, we introduce additional conditional information to make the learning process more controllable and flexible. Specifically, we use skill categories as conditional inputs to guide the generator in producing specific motion sequences under different skill contexts, thereby enabling simultaneous learning of multiple skills. This method not only allows handling multiple tasks within the same network but also ensures that the generation and learning of each task or skill can be adjusted according to the different requirements of the conditional variables, thus avoiding the task interference problems that may occur in traditional approaches.

To further improve the quality and controllability of skill-conditioned motion generation, we extend our framework by introducing an additional skill discriminator. While the conditional generator enables the policy to produce diverse motion sequences under different skill contexts, it does not explicitly ensure that the generated behaviors match the characteristics of the intended skill. To address this, the skill discriminator is designed to classify the type of skill expressed 
by the current behavior, using reference samples from the expert dataset as supervision. By incorporating this discriminator, the agent receives informative feedback during training, 
guiding the policy to better align the generated motions with the desired skill characteristics and enhancing both motor capability and control precision.

\textbf{This mechanism enhances the policy's dynamic adaptability to diverse skill requirements while improving robust training stability and cross-skill generalization performance.} We validate the proposed approach through extensive experiments in both simulation and on a physical quadruped robot. The results demonstrate that our system is capable of generating diverse motion skills conditioned on user-specified commands, and achieve smooth transitions between skills during motion.

\begin{table}[h]
\caption{DYNAMIC PARAMETERS AND THE RANGE OF THEIR RANDOMIZATION VALUES USED DURING TRAINING}
\label{DYNAMIC_PARAMETERS}
\begin{center}
\begin{tabular}{c c c}
\toprule
Parameters & Range [Min, Max] & Unit\\
\hline
Link Mass & [0.8, 1.2] × nominal value & Kg\\
Payload Mass & [0, 3] & Kg\\
Payload Position & [-0.1, 0.1] relative to base origin & m\\
Ground Friction & [0.05, 1.75] & -\\
Motor Strength & [0.8, 1.2] × motor torque & Nm\\
Joint Kp & [0.8, 1.2] × 20 & -\\
Joint Kd & [0.8, 1.2] × 0.5 & -\\
Initial Joint Positions & [0.5, 1.5] × nominal value & rad\\
\bottomrule
\end{tabular}
\end{center}
\end{table}

\section{RELATED WORK}

The reward function plays a critical role in reinforcement learning (RL), as it directly affects the efficiency of learning, the performance of the final policy, and the overall success in solving the target task. Motion imitation has been shown to be an effective tool for simplifying reward design [6]. Currently, the primary approaches in this area can be broadly categorized into two types: trajectory-based methods and style-based methods.

Trajectory-based methods resemble traditional RL approaches, where the reward is computed based on the discrepancy between the agent’s current state and the corresponding state in a reference trajectory. This encourages the agent to produce motion sequences that closely follow the expert demonstrations [8]. Prior studies have successfully applied this approach to the learning of various motor skills, including locomotion, jumping, and backflipping [9][10].

Style-based approaches differ primarily in their treatment of temporal information. In trajectory imitation, the agent explicitly minimizes the difference between its motion and the reference trajectory at each timestep, resulting in time-aligned behavior replication. In contrast, style-based methods focus on matching the overall distribution or characteristics of the motion, without requiring strict temporal correspondence.

These methods typically employ a discriminator to distinguish between reproduced motions and reference motions, and imitation rewards are derived from the discriminator's output [11], enabling adversarial training. Since style-based approaches do not require frame-by-frame comparison, they 
offer greater flexibility and tolerance to temporal variations. However, when the dataset contains a wide range of diverse behaviors, it becomes challenging for the state-transition discriminator to learn a meaningful distinction across different skills, potentially leading to mode collapse.

Subsequent work [12] extended the AMP framework into a latent skill space, enabling the learning of large-scale, reusable adversarial skill embeddings for physically simulated agents. Furthermore, [13] demonstrated that AMP can be adapted to train quadruped robots to acquire agile motor skills 
using the Wasserstein GAN formulation and by transitioning from coarse, partially observed demonstrations. AMP-based methods have been further applied to quadrupedal locomotion on complex terrains, demonstrating strong and agile capabilities [25], [26].

In multi-task style-based reinforcement learning, robots often struggle to balance velocity tracking accuracy with the expression of complex skill behaviors. This imbalance can hinder the development of adaptive behaviors across varying environments and speed conditions. To mitigate task interference in multi-task settings, one approach is to adopt hierarchical training [12], [14], which decomposes the RL problem into a pretraining phase and a downstream task learning 
phase. This allows motion skills to be reconstructed from large-scale datasets. However, freezing the lower-level skill networks during fine-tuning limits their adaptability to new environments and tasks, thereby reducing the flexibility and scalability of the overall system.

An alternative is to leverage unsupervised skill discovery [16], which combines Generative Adversarial Networks (GANs) with unsupervised learning techniques to enable the policy to 
effectively imitate diverse skills from expert data. This approach also helps mitigate the mode collapse issue commonly observed when applying GANs to large-scale unlabeled datasets. However, its success heavily depends on the quality and diversity of the expert data. If the dataset lacks sufficient richness, the generator may fail to learn useful and discriminative skills.

Another line of work introduces the Multi-AMP framework [17] for multi-skill policy learning, where each expert dataset is paired with a dedicated adversarial network, enabling more complex skill transfer and deployment. Nonetheless, as the number of skills increases, this method incurs significant computational and memory overhead and often fails to learn smooth transitions between different skills.

\begin{table}[h]
\centering
\caption{Reward function parameters}
\begin{tabular}{ccc}
\toprule
\textbf{Term} & \textbf{Equation} & \textbf{Weight} \\
\midrule
\multirow{2}{*}{$r^{\text{task}}$} & $\exp(-\|v_{xy}^{cmd}-v_{xy}\|_{2}/0.15)$ & 1.5 \\
                                   & $\exp(-\|\omega_{z}^{cmd}-\omega_{z}\|_{2}/0.15)$ & 0.75 \\
\midrule
$r^{\text{style}}$ & (4) & 1.0 \\
\hline
$r^{\text{skill}}$ & (6) & 0.3 \\
\bottomrule
\end{tabular}
\label{tab:reward_params}
\end{table}

\begin{figure*}[t]  
    \centering
    \framebox{\parbox{6.8in}{
    \includegraphics[width=0.979\textwidth, trim=2 2 2 2, clip]{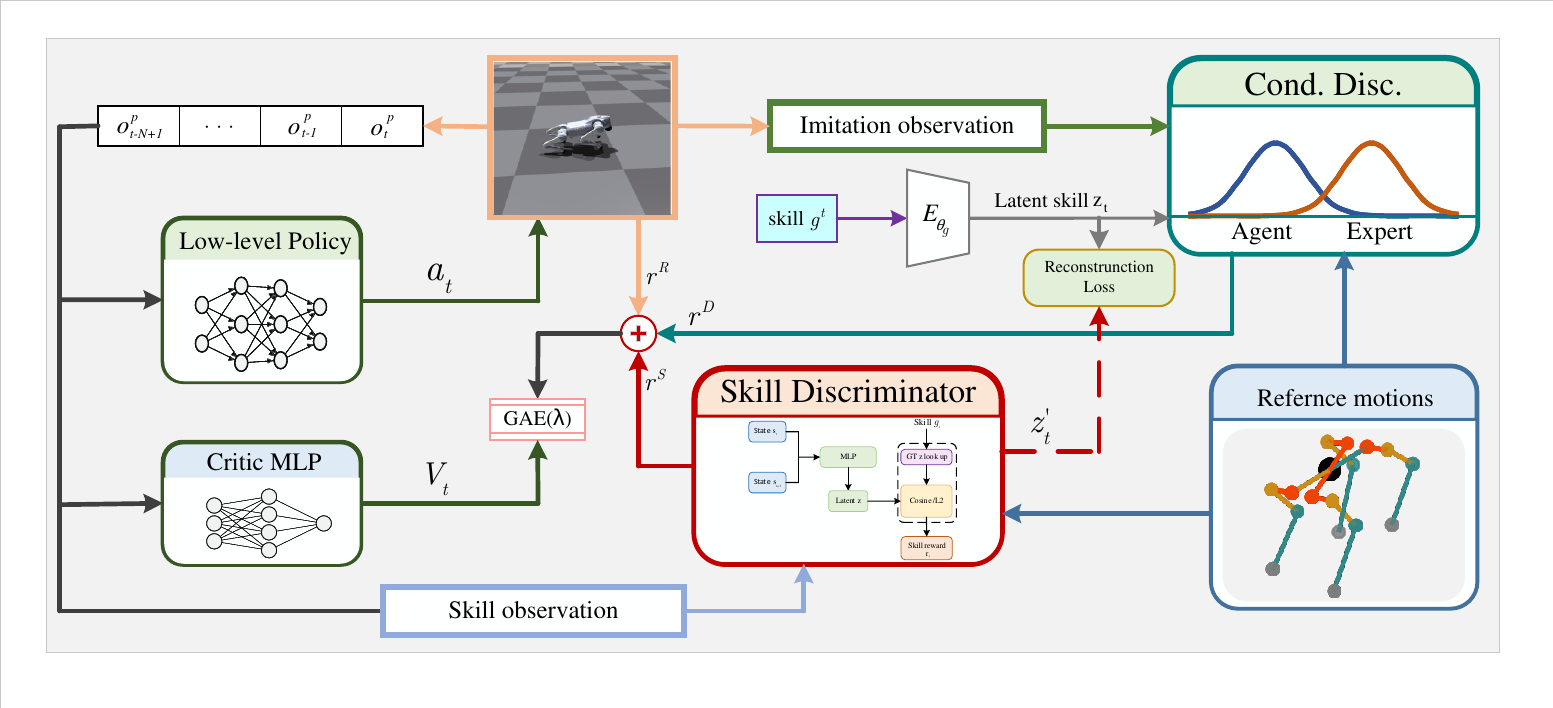}
    \vspace{-30pt} 
    \caption{Overview of the training framework. We adopt an asymmetric actor-critic structure, where the policy is guided by a conditional discriminator to learn natural gait skills from the expert dataset. A skill discriminator is used to construct a latent skill space, enabling the agent to reconstruct multiple gait skills as needed.}
    \label{fig:framework}
    }}
\end{figure*}

\section{METHOD}

In this section, we describe our approach, which enables the learning of diverse and distinguishable motion skills directly from data. Our method enables a single policy network 
to generate stylistically distinct motion behaviors by conditioning on learned skill representations. Unlike conventional approaches that rely on task-specific networks or manual 
decomposition, our unified framework simultaneous learning of multiple motor skills in a single model, thereby significantly enhancing both scalability and generalization capability.

\subsection{Reinforcement Learning Problem Formulation} 

Reinforcement learning (RL) in quadruped locomotion can be formulated as a Partially Observable Markov Decision Process (POMDP), defined by the tuple (\textit{S,O,A,R,p,r}). The agent interacts with the environment according to a policy in order to optimize a predefined objective function [18]. The objective is to learn a policy that maximizes the expected discounted cumulative return \(J(\pi)\),

\begin{equation}
J(\pi)=E_{p(\tau \mid \pi)}\left[\sum_{t=0}^{T-1} \gamma^{t} r_{t}\right]
\end{equation}

where $\gamma^t \in [0, 1)$ is a discount factor.

To train our multi-gait quadruped locomotion controller, we leverage the Proximal Policy Optimization (PPO) algorithm for policy optimization. Our framework adopts an asymmetric architecture and incorporates a self-supervised learning paradigm to train the gait discriminator, enabling effective distinction between different locomotion patterns.

\textbf{State Space:} In our framework, the actor network processes only simple observations ${{o}_{t}}\in {{\mathbb{R}}^{48}}$ including: (1)the desired angular velocity $\omega$, (2)the desired skill vector $g^t$ and desired velocity command $c_{t}=\left(v_{x}^{c m d}, v_{y}^{c m d}, \omega_{z}^{c m d}\right) \in \mathbb{R}^{3}$, (3)proprioceptive data from joint encoders $\theta_t$ and $\dot{\theta_t }$, (4)the gravity projection, (5) the network output from the previous timestep $a_{t-1}$. The is designed as a one-hot vector to avoid introducing implicit ordinal biases. The desired velocity command $c_t$ represents the target longitudinal, lateral, and yaw angular velocities in the robot’s body coordinate frame.

\textbf{Action space:} The policy outputs $\text{action} \in \mathbb{R}^{12}$ as a joint position offset, and the target joint position is defined as $\theta_{\text{target}} = \theta_{\text{init}} + k a_{t}$. It is then sent to the low-level PD controller to compute the target torque. To ensure that small deviations during motion can be quickly corrected, the calf joint motors usually require a higher proportional gain to increase the feedback strength. Therefore, we set our $K_{p\text{-calf}} = 40$, while all other joints use $K_{p} = 30$, $K_{d} = 1$.

\textbf{Reward Function}: The objective is to construct a concise yet effective reward function that incentivizes the agent to acquire robust and agile legged locomotion with natural gait patterns. The overall reward is composed of task reward $r^{task}$, style reward $r^{style}$, and skill reward $r^{skill}$:

\begin{equation}
r_{t} = \omega^{\text{task}} r_{t}^{\text{task}} + \omega^{\text{style}} r_{t}^{\text{style}} + \omega^{\text{skill}} r_{t}^{\text{skill}}
\end{equation}

where $\omega$ denotes the weighting coefficient assigned to each individual reward component. A detailed specification of the reward terms is provided in Table~II. Task rewards $r^{\text{task}}$ are primarily used to drive the robot to achieve pre-defined motion objectives, such as maintaining balance, forward velocity, or turning accuracy. Style rewards $r^{\text{style}}$ encourage the agent to generate more natural motion styles by comparing features from expert datasets (e.g., joint angles, foot-end positions). Skill rewards $r^{\text{skill}}$, based on a skill discriminator, incentivize the agent to learn diverse and distinctly identifiable skill behaviors.

Our method learns multi-gait and multi-frequency locomotion skills simultaneously through a single policy network, without the need for phased training or external switching mechanisms. In our framework, the conditional vector is encoded into skill latent variables $z_t$ through an encoder module $E_\theta$. A conditional AMP discriminator is employed to learn distinctive features from different skill data $(s_t, s_{t+1}, z_t)$, while a skill discriminator reconstructs the skill latent variables $z_t$ based on the observed data.

\begin{figure*}[t]
    \framebox{\parbox{6.8in}{
    \centering
    \includegraphics[width=0.96\linewidth]{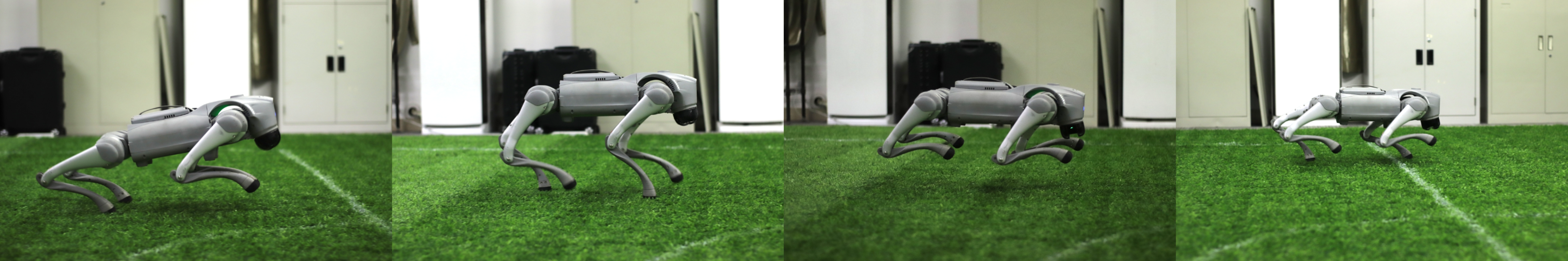}
    \caption{Skill transition performance of the policy in the real-world environment.}
    \label{fig:motion_transition}
    }}
\end{figure*}

\subsection{Motion Dataset Generation}
We generate quadrupedal locomotion sequences on flat terrain using a model-based control approach, without relying on real-world motion capture data. This method not only significantly reduces data collection costs but also eliminates the need for additional motion retargeting techniques.

Our dataset comprises various locomotion trajectories, i.e., forward, backward, leftward, rightward, left-turning, right-turning, and composite motion trajectories. It encompasses multiple gait patterns such as trot, pace, bound, and pronk, with each trajectory lasting approximately 4 seconds.

The motion trajectory data includes body position, body orientation (quaternion), joint positions, joint velocities, foot positions, and foot velocities. 

To improve training efficiency and enhance model stability during the initial phase, we introduce a state transition preloading strategy in the data loading process. Specifically, multiple state transition pairs (i.e., the current state and its next state) are uniformly sampled from expert demonstration data and preloaded into memory prior to training initialization. This strategy effectively reduces data loading time during early training and provides sufficient positive samples for the discriminator. Additionally, each trajectory is associated with its corresponding gait feature label, enabling simultaneous facilitate multi-gait classification and control tasks.

\subsection{Learning Diverse Locomotion Skills}

Our skill learning framework extends the Adversarial Motion Prior (AMP) approach [19], which has demonstrated effectiveness in acquiring motion styles from expert demonstrations. However, extending this method to accommodate a wide range of locomotion behaviors remains challenging, as AMP tends to suffer from mode collapse when applied to multi-behavior datasets. To address this limitation, inspired by [20], we propose an improved approach based on a conditional discriminator architecture. An overview of our method is illustrated in Fig 1.

Specifically, our conditional discriminator uses skill embeddings to separate locomotion types (walking, running, jumping), simultaneously verifying sample authenticity and skill adherence. This design mitigates mode collapse and enhances the diversity of learned behaviors.

Formally, the conditional discriminator $D_{\theta}$ is a parameterized neural network that predicts whether a given state transition ($s_{t}, s_{t+1}$), conditioned on a skill latent variable $z^{p}$, originates from the expert dataset or from the learned policy. Each state $s_{t}^{\text{CAMP}} \in \mathbb{R}^{43}$ comprises joint positions, joint velocities, base linear velocity, base angular velocity, base height, and foot positions.

The training objective is defined as follows:
\begin{equation}
\begin{aligned}
\underset{\theta}{\arg\min} \, & \mathbb{E}_{d^{M}}\left[\left(D_{\theta}(s_{t}, s_{t+1} \mid z^{p}) - 1\right)^{2}\right] \\
& + \mathbb{E}_{d^{\pi}}\left[\left(D_{\theta}(s_{t}, s_{t+1} \mid z^{p}) + 1\right)^{2}\right] \\
& + \omega_{gp} \, \mathbb{E}_{d^{M}}\left[\left\| \nabla_{\theta} D_{\theta}(s_{t}, s_{t+1} \mid z^{p}) \right\|^{2}\right]
\end{aligned}
\end{equation}

The first two terms of the training objective follow the Least Squares GAN (LSGAN) formulation, where the discriminator is encouraged to output values close to 1 for real samples and close to $-1$ for generated samples. The last term is a gradient penalty term with coefficient $\omega_{gp}$, which penalizes non-zero gradients on expert samples, thereby improving training stability [21]. The reward function for training the policy is defined as follows:
\begin{equation}
r^{\text{style}} = \max \left[ 0, 1 - 0.25 \left( D_{\theta} \left( s_{t}, s_{t+1} \mid z^{p} \right) - 1 \right)^{2} \right]
\end{equation}

Through this condition-based expert data learning approach, a single policy network can simultaneously learn multi-gait and multi-frequency locomotion skills without phased training or external switching mechanisms, enabling the robot to switch between different gait skills during motion without compromising stability.

\subsection{Multi-Skill Reconstruction}

Although the style reward can implicitly guide the agent to learn motion styles from expert datasets, it provides relatively soft priors and may fail to offer sufficiently precise feedback in fine-grained or task-specific control scenarios. As a result, the agent may struggle to reuse the learned motion skills effectively. Inspired by [22], we augment the actor network with both the agent's state observations $o_{t}^{p}$ and a skill condition vector $g^{t} = (c_{1}, c_{2}, \ldots, c_{l})$, where each dimension represents an independent skill and is constrained to lie within the range $[0, 1]$. We further design a skill discriminator $D_{\text{skill}}$, which aims to learn a mapping from state transition pairs $[s_{t}, s_{t+1}]$ to a latent skill embedding $z \in \mathbb{R}^{d_{z}}$ in the skill style space.

During the training phase, to guide the policy towards learning state transitions that are consistent with the target skill, we design a similarity-based reward mechanism. This mechanism encourages the predicted skill latent variable $z_{t}^{'}$, output by the skill discriminator $D_{\text{skill}}$, to closely approximate the skill embedding $z_{t}$ provided by the conditional discriminator. Specifically, the $z_{t}$ vector is obtained from the skill embedding layer within the conditional discriminator.
\begin{figure}
    \centering
    \includegraphics[width=1.01\linewidth]{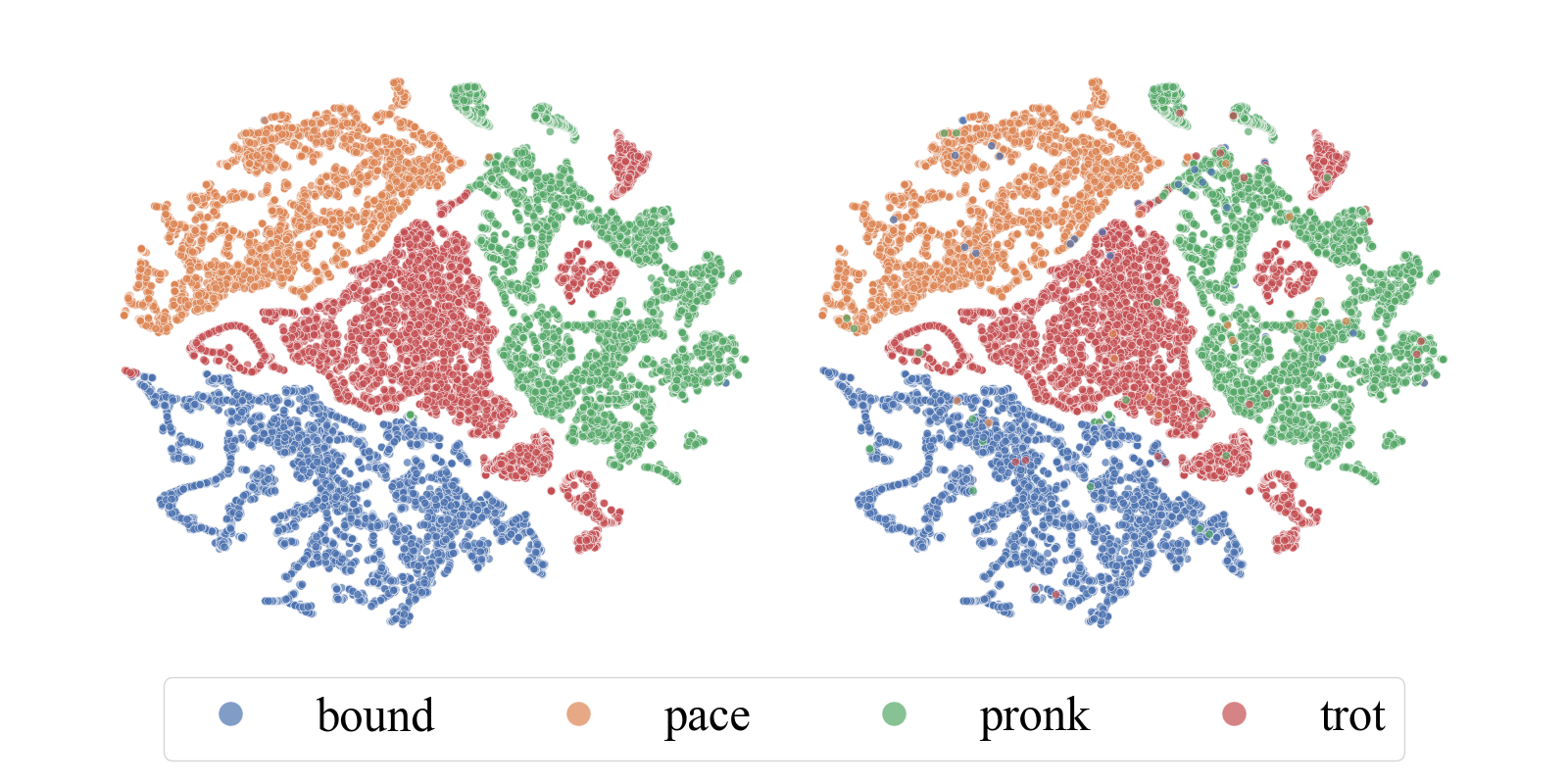}
    \caption{The right plot shows the predicted skill clusters by the skill discriminator, while the left plot presents the ground-truth skill labels.}
    \label{fig:t-sne}
\end{figure}
\begin{equation}
z = \mathrm{E}(y), \quad \hat{z} = f_{\theta}(s_{t}, s_{t+1})
\end{equation}

where $E$ denotes the skill embedding layer shared from the conditional AMP discriminator; $f_{\theta}$ represents the forward network of the skill discriminator; and $y$ is the discrete skill 
label associated with the expert data.

The skill reward is computed as the cosine similarity between the predicted skill latent variable and the target skill embedding
\begin{equation}
r^{\text{skill}} = \cos(z', z) = \frac{z' \cdot z}{\|z'\| \, \|z\|}
\end{equation}
e.q. 6 measures the consistency between the state transitions generated by the policy and the target skill pattern. Meanwhile, the skill discriminator is updated using expert data, with the training objective defined as follows:
\begin{equation}
\mathcal{L}_{\text{expert}} = \mathbb{E}_{(s_t, s_{t+1}, y) \sim D_{\text{exp}}} \left[ \|\hat{z} - z\|_{2}^{2} \right]
\end{equation}
where this loss function can be viewed as a supervision signal for the style recognition capability of the discriminator, encouraging it to map state transitions corresponding to different skills into their respective regions in the skill style space. To enhance training stability and prevent discriminator overfitting, we introduce a gradient penalty term:
\begin{equation}
\mathcal{L}_{\mathrm{GP}} = \lambda \cdot \mathbb{E}_{(s_{t}, s_{t+1})} \left[ \left\| \nabla_{x} f_{\theta}(s_{t}, s_{t+1}) \right\|_{2}^{2} \right]
\end{equation}
\vspace{5pt}

\section{EXPERIMENTS}

In this section, our experimental evaluation encompasses simulation tests, real-world validation on the Unitree Go2 quadruped, and ablation studies to verify key design choices.

\textbf{Simulation:} We trained 4096 parallel agents using the Isaac Gym simulator [23]. The policy was trained with million simulated time steps. In simulation, the policy operates at a control frequency of 50 Hz. Training is conducted on an NVIDIA RTX 4060Ti GPU and takes approximately 7 hours. Policy robustness is evaluated via cross-simulator transfer tests, including Gazebo and MuJoCo environments.
\begin{figure}
    \centering
    \includegraphics[width=1.01\linewidth]{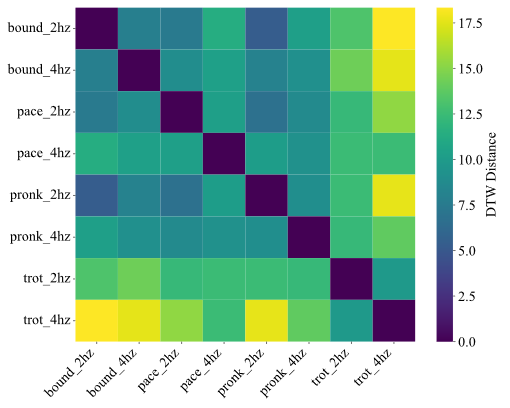}
    \caption{DTW distances of gait skills in the latent space. Darker colors indicate higher proximity between two gait skills in the latent space.}
    \label{fig:DTW}
\end{figure}
To bridge the sim-to-real gap and account for the non-stationarity of robot dynamics and system uncertainties, we implement domain randomization techniques during training [24]. Specifically, we randomize robot hardware parameters, control parameters, and environment parameters within predefined ranges to make the simulated environment more representative of the real world. Details of the randomized parameters are provided in Table I.

\subsection{Skill Extraction and Prediction}

To evaluate the model's ability to distinguish between multiple gait skills, we design a visualization experiment that combines t-SNE dimensionality reduction with K-means clustering. This experimental setup enables quantitative analysis of gait skill separability in the learned latent space. We select four representative quadrupedal gait skills: bounding, pacing, pronking, and trotting, and perform the analysis on the reference data corresponding to each skill.

Specifically, motion reference data (10s segments) were acquired for each gait skill, containing complete state observations (positions, velocities and so on), and extracted feature representations using the trained skill discriminator. To eliminate the influence of scale differences across feature 
dimensions, all extracted features were standardized. 

Subsequently, K-means clustering was performed on the standardized features, and the clustering results were compared with the ground-truth gait labels. Finally, we applied t-SNE to reduce the high-dimensional features to a two-dimensional space and plotted the distributions of both the clustering labels and the ground-truth labels, as shown in Figure 3. These visualizations provide an intuitive illustration of the model’s clustering performance and the decision boundaries in skill discrimination.

\begin{table}[h]
\centering
\caption{NETWORK ARCHITECTURE}
\label{tab:network_architecture}
\begin{tabular}{cccc}
\toprule
\textbf{Module} & \textbf{Inputs} & \textbf{Hidden Layers} & \textbf{Outputs} \\
\midrule
Actor (MLP) & $o_{t}$ & [512,256,128] & $a_{t}$ \\
Critic (MLP) & $o_{t}, x_{t}$ & [512,256,128] & $V_{t}$ \\
$D_{\theta}$ & $s_{t}, s_{t+1}$ & [1024,512] & $d_{t}$ \\
$D_{\text{skill}}$ & $s_{t}, s_{t+1}$ & [512,256] & $p_{t}$ \\
\bottomrule
\end{tabular}
\end{table}

To measure the similarity and reusability between gait skills, we adopt Dynamic Time Warping (DTW) to calculate the distance between the latent representations of different gait skills at different frequencies. As shown in Fig.~4, the dark blocks along the diagonal region (e.g., {trot\_2Hz} and {trot\_4Hz}) with DTW $\approx 3.2$ indicate that the same type of gait maintains high latent similarity under cross-frequency conditions, verifying the robustness of skill representation under frequency perturbations.

Our analysis reveals the DTW distance between {pronk\_2Hz} and {bound\_2Hz} is significantly smaller than that between {bound\_2Hz} and {bound\_4Hz}, indicating that different gaits at similar frequencies may exhibit higher temporal similarity due to convergence of locomotion rhythm.

\subsection{Evaluation Experiments of the Learning Framework}
To demonstrate the effectiveness of each module within the framework, we conducted ablation studies on different components and compared our method with other controllers. We compared our approach with the following methods:

\begin{itemize}
    \item \textbf{Baseline:} a framework that uses AMP to train quadruped locomotion.
    
    \item \textbf{Ours w/o skill observation:} our framework without skill observation ablation.
    
    \item \textbf{Ours w/o skill conditioning:} ablating skill conditioning, using only the standard AMP discriminator.
    
    \item \textbf{Ours w/o skill discriminator:} directly removing the skill discriminator model.
\end{itemize}

For a fair comparison, all RL methods use the same low-level network architecture shown in Table~III. Each policy is trained for 15,000 steps, and all randomization settings remain consistent during training. In the testing phase, we evaluate the methods based on their skill reproduction ability, with detailed results shown in Table~IV.

\begin{figure}[h]
    \centering
    \includegraphics[width=1.01\linewidth]{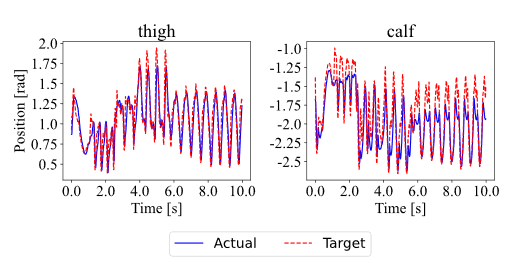}
    \caption{Comparison between target and actual joint positions of each motor in the FL leg. The joint positions demonstrate good tracking accuracy during motion.}
    \label{fig:joint_pos}
\end{figure}
The experimental results show that when the skill observation $g^{t}$ is removed from the policy input, the model struggles to distinguish behavioral differences between skills. The policy tends to generate only one dominant skill and lacks the ability to switch skills according to commands or task requirements. When skill condition $z^{p}$ is removed, the policy can still learn behaviors with natural motion; however, its ability to differentiate between different skills significantly decreases. This is reflected in the generated gaits, which tend to be ambiguous or blended, lacking clear skill structures and exhibiting obvious ``mode collapse."

\begin{table}[t]
\centering
\caption{COMPARISON OF ABLATION EXPERIMENT RESULTS}
\label{tab:COMPARISON OF ABLATION EXPERIMENT RESULTS}
\begin{tabular}{lccc}
\toprule
\textbf{Benchmarks} & \textbf{Multiple gaits} & \textbf{Gait control switch} \\
\midrule
Baseline & No & No \\
Ours & Yes & Yes \\
Ours w/o $g^{t}$ & Yes & No \\
Ours w/o $z^{p}$ & No & No \\
Ours w/o $r^{\text{skill}}$ & Yes & No \\
\bottomrule
\end{tabular}
\end{table}

When the skill discriminator model is removed, experimental observations indicate that if the expert dataset contains only two distinct skills, the policy can still effectively distinguish and reproduce the target skill based on the input condition. However, when the number of skills exceeds two, the 
generated behaviors show signs of skill confusion or neutrality, failing to accurately reproduce the style of a specific skill. This demonstrates that the skill discriminator plays a critical role in 
multi-skill learning by enhancing the expressiveness of the policy and improving skill disentanglement.

\subsection{Stable Skill Transitions}

In this section, we systematically evaluate the overall performance of the proposed control method by analyzing tracking accuracy and the quality of skill transitions. Fig. 2. illustrates the skill transition experiment conducted on the hardware platform. In this experiment, the robot autonomously generates and switches among multiple gait skills in response to user commands based on the learned policy. The results show that, with the aid of our designed controller, the robot can achieve natural and smooth gait transitions.

In addition, we designed a 10-second continuous skill switching sequence. Figure 5 shows the joint trajectory tracking results compared with the reference trajectory, and Fig. 6. shows the contact sequence of both feet to quantitatively evaluate the controller's action execution ability.

As shown in Fig.~5, the position trajectory tracking of the thigh and calf motors on the RL leg is illustrated, where the blue lines indicate the actual joint positions and the red lines indicate the target trajectories. The results show that the average tracking accuracy of the joint trajectories reaches 91.23\% and 91.38\%, respectively, indicating that the control policy possesses strong execution performance and tracking capability.

\begin{figure}[t]
    \centering
    \includegraphics[width=1.01\linewidth]{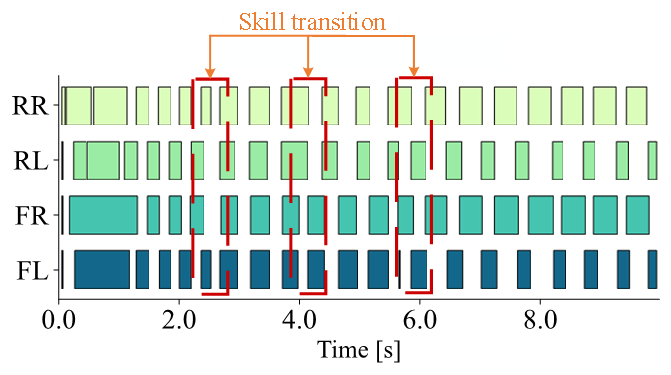}
    \caption{Foot contact sequence results of the robot. FL, FR, RL, and RR represent the front-left, front-right, rear-left, and rear-right legs, respectively.}
    \label{fig:phase_diagram}
\end{figure}

Fig.~6 presents the foot phase diagram during the multi-skill switching process, providing an intuitive visualization of continuity and stability between gaits. It can be observed that the proposed method enables smooth and natural transitions between different gait skills. The entire process occurs without noticeable oscillations or instability, and the previous motion behaviors remain unaffected.

\subsection{Sim-to-real Transfer}

We deploy the trained policy on a real Unitree Go2 robot, as shown in Fig.~2. The deployment results demonstrate smooth transitions between different gait patterns. Notably, this process does not require additional parameter tuning and can be directly deployed on other quadruped robots with similar specifications, such as the Unitree A1 robot.

\section{CONCLUSION}

Our research demonstrates that our framework effectively integrates diverse expert data, addressing the critical challenge of seamlessly learning and transitioning between multiple skills using a single policy, while overcoming the limitations of fragmented skill execution and rigid behavior 
switching in previous approaches. The results demonstrate that our skill discriminator effectively addresses the challenge of precise skill reconstruction by constructing skill reward functions. Unlike previous algorithms that relied on heuristic reward engineering or suffered from ambiguous skill representations, our discriminator explicitly guides policy learning to recover and distinguish expert skill data, thereby ensuring high-fidelity imitation and robust transitions. Moreover, this approach can be directly deployed on real robotic systems without additional parameter tuning, exhibiting strong transferability and system robustness.

Although the experimental results in this study primarily focus on quadruped robots, the proposed control method is not limited to quadrupedal morphology and does not rely on any specific motion generation tools. In future work, this method can be extended to humanoid robots to achieve more 
complex skill behaviors.





\nocite{*}
\bibliographystyle{ieeetr}
\bibliography{Mybib}


\end{document}